\documentclass{article}

\usepackage[accepted]{icml2026}

\makeatletter
\renewcommand{\Notice@String}{}
\makeatother

\usepackage{times}
\usepackage{hyperref}
\usepackage{url}
\usepackage{booktabs}
\usepackage{amsfonts}
\usepackage{amsmath}
\usepackage{nicefrac}
\usepackage{microtype}

\usepackage{graphicx}
\usepackage{subcaption}

\icmltitlerunning{Look-Ahead-Bench: Standardized Benchmark of Lookahead Bias in PiT LLMs}

\begin{document}

\twocolumn[
\icmltitle{Look-Ahead-Bench: a Standardized Benchmark of Look-ahead Bias in Point-in-Time LLMs for Finance}

\icmlsetsymbol{equal}{*}

\begin{icmlauthorlist}
\icmlauthor{Mostapha Benhenda}{equal}
\end{icmlauthorlist}

\icmlcorrespondingauthor{Mostapha Benhenda}{\nolinkurl{mostaphabenhenda@gmail.com}}


\icmlkeywords{ Large Language Models, Lookahead Bias, Point-in-Time Models, Financial Benchmarking, Backtesting}

\vskip 0.3in
]

\printAffiliationsAndNotice{}

\begin{abstract}
We introduce Look-Ahead-Bench, a standardized benchmark measuring look-ahead bias in Point-in-Time (PiT) Large Language Models (LLMs) within realistic and practical financial workflows. Unlike most existing approaches that primarily test inner lookahead knowledge via Q\&A, our benchmark evaluates model behavior in practical scenarios. To distinguish genuine predictive capability from memorization-based performance, we analyze performance decay across temporally distinct market regimes, incorporating several quantitative baselines to establish performance thresholds.

 We evaluate prominent open-source LLMs---Llama 3.1 (8B and 70B) and DeepSeek 3.2---against a family of Point-in-Time LLMs (Pitinf-Small, Pitinf-Medium, and frontier-level model Pitinf-Large) from PiT-Inference. Results reveal significant lookahead bias in standard LLMs, as measured with alpha decay, unlike Pitinf models, which demonstrate improved generalization and reasoning abilities as they scale in size. This work establishes a foundation for the standardized evaluation of temporal bias in financial LLMs and provides a practical framework for identifying models suitable for real-world deployment. Code is available on GitHub: \url{https://github.com/benstaf/lookaheadbench}
\end{abstract}

\section{Introduction}
\label{sec:introduction}

Large Language Models are revolutionizing quantitative finance. FinGPT \citep{yang2023fingpt},  BloombergGPT \citep{wu2023bloomberg}, FinMem \citep{yu2023finmem}, FinRL-DeepSeek \citep{benhenda2025finrl}, and TradingAgents \citep{xiao2025trading}, along with multi-agent architectures like Hedge-Agents \citep{li2025hedge} and the AI Hedge Fund \citep{virattt2024hedge}, have demonstrated remarkable capabilities in processing financial data, generating trading signals, and executing complex investment strategies. However, beneath these promising results lies a fundamental methodological challenge that threatens the validity of LLM-based financial applications: look-ahead bias.

Look-ahead bias occurs when models access information that would not have been available at the time of prediction, creating artificially inflated performance metrics that evaporate in real-world deployment. Consequently, backtested returns often collapse once the model's knowledge window ends and trading gets into genuinely unknown territory \citep{li2025profit}.

\subsection{The Challenge of Temporal Bias in Financial LLMs}

LLMs are pre-trained on web-scale corpora containing extensive financial data, including post-hoc explanations of market events, historical price movements, and retrospective analyses. When an LLM encounters a prompt about ``NVIDIA's performance in 2023,'' it may have been trained on text explicitly stating ``NVIDIA surged 190\% in 2023 on AI boom'' \citep{li2025profit}. In this case, the model does not learn predictive relationships; it memorizes outcomes and recites them during evaluation.

This pre-training contamination creates systematic biases that are particularly pernicious in finance, where temporal causality is paramount. Several studies have documented various manifestations of this issue:

\begin{itemize}
\item \textbf{Training Data Leakage:} Models recall specific stock prices, earnings figures, and market events directly from their training corpus \citep{lopez2025memorization}.
\item \textbf{Entity Memorization:} Even with masked identifiers, models can recognize companies from contextual clues and retrieve associated historical performance \citep{sarkar2024stories}.
\item \textbf{Temporal Reasoning Failures:} Models struggle to distinguish between information available at different time points, leading to logically impossible predictions \citep{yan2025temporal,gao2025test}.
\end{itemize}

\subsection{Limitations of Current Evaluation Approaches}

Existing approaches to detecting lookahead bias have focused primarily on testing inner knowledge through question-answering tasks, or evaluating the memorization of specific facts. While valuable, these evaluation methods may not reflect the practical consequences of bias in real-world financial settings.

The pragmatic test of a financial model's capability does not lie in its ability (or lack thereof) to recall historical patterns from its training data, but in its practical capacity to generalize to genuinely novel market conditions.

\subsection{Our Contributions}

We introduce Look-Ahead-Bench, a standardized benchmark that addresses these limitations through three key contributions:

\begin{enumerate}
\item \textbf{Practical Trading Workflow Integration:} Unlike knowledge-based benchmarks, our evaluation uses realistic agentic trading systems that make actual portfolio decisions. We utilize the AI Hedge Fund framework \citep{virattt2024hedge}, a widely adopted open-source project (+45k stars on GitHub), ensuring evaluation moves beyond toy problems to assess real-world deployment viability.

\item \textbf{Point-In-Time LLMs Evaluation:} We explicitly evaluate Pitinf models---commercial LLMs specifically designed to effectively remove lookahead bias---alongside standard foundation models.

\item \textbf{Alpha Decay Metric:} We introduce a rigorous metric for measuring bias by quantifying the performance drop between in-sample (training window) and out-of-sample (future) periods.
\end{enumerate}

Our results demonstrate that standard LLMs exhibit significant lookahead bias, with alpha decay exceeding -15 percentage points between periods, while Pitinf models maintain stable performance. This work provides a diagnostic tool for identifying bias in financial LLMs.

\section{Related Work}
\label{sec:related}

\subsection{Lookahead Bias in Financial LLMs}

The recognition of lookahead bias as a critical issue in financial LLM applications has grown rapidly in recent years. \citet{glasserman2023assessing} provided one of the earliest systematic assessments, investigating both lookahead bias and distraction effects in GPT-based sentiment analysis for stock return prediction. Their work revealed that anonymizing company identifiers could actually improve performance by reducing distraction effects, while highlighting the challenge of disentangling genuine sentiment analysis from memorized associations.

\citet{levy2024lookahead} emphasized implications for financial applications and proposed mitigation strategies including careful data cutoff management and entity embedding neutralization. \citet{gao2025test} developed a statistical test for detecting lookahead bias using Lookahead Propensity (LAP), a measure based on membership inference attack techniques. Their approach correlates model familiarity with training data (estimated through token probability analysis) with forecast accuracy. While innovative, this method focuses on individual prediction tasks rather than end-to-end trading system evaluation.

Other contributions have expanded both the empirical and methodological study of lookahead bias in financial language models, proposing standardized diagnostics, benchmark-style evaluations, and controlled temporal stress tests that reveal performance inflation caused by future data contamination in forecasting and trading settings \citep{noguer2024lookahead,lopez2025contamination}.

\subsection{Information Leakage and Memorization}

The broader issue of information leakage in LLMs has received extensive attention. \citet{li2025profit} introduced FinLeak-Bench and the FactFin framework to systematically quantify information leakage across four dimensions, demonstrating that most published LLM-based agents fail to beat baselines once their knowledge cutoff is passed.

\citet{lopez2025memorization} documented the ``memorization problem'' through extensive testing of GPT-4o's recall capabilities, showing that models can recall exact S\&P 500 closing prices with less than 1\% error for dates within their training window, while errors ``explode'' for post-cutoff dates. \citet{carlini2021extracting,carlini2022quantifying} established theoretical and empirical foundations for understanding memorization in large language models, demonstrating that training data can be extracted from models through carefully designed queries.

Complementary studies further show that pretraining contamination and implicit temporal leakage can persist even without explicit date cues, and that leakage effects scale with model size unless explicitly mitigated \citep{kong2025fusing,drinkall2024time}.

\subsection{Point-in-Time and Temporally Aware Models}

The development of Point-in-Time (PiT) models represents a direct response to lookahead bias concerns. Models like Time Machine GPT \citep{drinkall2024time}, ChronoGPT \citep{he2025chrono} and DatedGPT \citep{yan2025temporal} introduce time-aware frameworks that train models strictly on pre-cutoff data to ensure temporal integrity. Their approach demonstrates that explicit temporal constraints can prevent future information leakage. \citet{merchant2025fast} proposed logit adjustment from specialized auxiliary models, allowing both verbatim and semantic knowledge to be removed without retraining large base models.

There are also proprietary Point-In-Time LLMs designed for backtesting, such as those from PiT-Inference \citep{pitinference2026}. These are available in three sizes: Small size ($<$10B parameters) for low-latency, Medium size ($<$100B parameters), and Large size ($>$500B parameters) for frontier-grade reasoning tasks.

\subsection{Agentic Trading Systems}

The application of LLMs to agentic trading has produced numerous innovative systems. \citet{yu2023finmem,yu2024fincon} developed FinMem and FinCon, demonstrating sophisticated memory and multi-agent capabilities. \citet{zhang2024finagent} created a multimodal foundation agent with tool augmentation, while \citet{huang2024finrobot} introduced FinRobot for equity research and valuation.

The AI Hedge Fund open-source project \citep{virattt2024hedge} provides LLM-based trading agents and serves as the foundation for our benchmark implementation, due to its significant impact (+45k stars on GitHub) on the community.

\section{Methodology}
\label{sec:methodology}

\subsection{Dual-Period Design}

Our evaluation compares model performance across two carefully selected periods to test models with knowledge cutoffs through 2023-2024, while maintaining similar market characteristics for fair comparison, like in FinLeak-Bench \citep{li2025profit}:

\paragraph{Period P1 (In-Sample):} April 1, 2021 -- September 30, 2021
\begin{itemize}
\item Duration: 6 months
\item Buy-and-Hold Return: +25.32\%
\item Purpose: Establish baseline performance within the potential training window of Llama 3.1 (8B, 70B) and DeepSeek 3.2.
\end{itemize}

\paragraph{Period P2 (Out-of-Sample):} July 1, 2024 -- December 31, 2024
\begin{itemize}
\item Duration: 6 months
\item Market Regime: AI-driven rally with mixed sentiment.
\item Buy-and-Hold Return: +24.75\% (similar to P1).
\item Purpose: Test generalization to post-cutoff periods for Llama 3.1 (8B, 70B) and DeepSeek 3.2.
\end{itemize}

\subsection{Portfolio and Trading Universe}

Our benchmark employs a focused trading universe consisting of five large-cap technology stocks:
AAPL (Apple Inc.), MSFT (Microsoft Corporation), GOOGL (Alphabet Inc.), NVDA (NVIDIA Corporation), and TSLA (Tesla Inc.).

\subsection{Evaluation Metrics}

\subsubsection{Alpha Calculation}

For each model and period, we calculate alpha as:
\begin{equation}
\alpha = R_{\text{LLM}} - R_{\text{Buy\&Hold}}
\end{equation}

This measures the excess return generated by the LLM strategy relative to a passive buy-and-hold benchmark.

\subsubsection{Alpha Decay Analysis}

The core bias indicator is alpha decay between periods:
\begin{equation}
\alpha_{\text{Decay}} = \alpha_{P2} - \alpha_{P1}
\end{equation}

Negative alpha decay indicates that the model's relative performance deteriorated in the out-of-sample period, suggesting potential lookahead bias in the in-sample results.

\subsection{Quantitative Baselines}
\label{subsec:quant-baselines}

We benchmark the AI agents against a suite of traditional quantitative strategies. These strategies span passive, systematic, trend-following, and contrarian approaches, providing a comprehensive reference for evaluating relative performance and robustness across market regimes. All strategies are executed with the same initial capital, fractional shares allowed, and monthly rebalancing (except where noted).

\subsubsection{Buy \& Hold (Passive)}
The simplest baseline allocates capital equally across all tickers at the start of the period and holds the positions without any rebalancing. This represents a passive, long-only exposure to the selected universe and serves as the reference for computing alpha (excess return over this benchmark).

\subsubsection{Equal-Weight Monthly Rebalance (Systematic)}
This strategy maintains equal dollar allocation to each ticker by rebalancing at the start of each month. It captures broad market exposure while controlling concentration risk, often outperforming naive buy-and-hold in diversified universes due to the rebalancing premium.

\subsubsection{Momentum (3-Month, Top Half)}
A classic trend-following strategy that, at each monthly rebalance, ranks tickers by their past 3-month total return and allocates equally to the top half (or all tickers equally if insufficient history). This exploits persistence in intermediate-term price trends and is one of the most robust quantitative factors historically.

\subsubsection{Mean Reversion (3-Month, Bottom Half)}
The contrarian counterpart to momentum: at each monthly rebalance, the strategy ranks tickers by past 3-month return and allocates equally to the bottom half of performers (or all equally if insufficient history). It bets on price reversals and tends to perform well in range-bound or high-dispersion regimes but can suffer in strong trends.

\subsubsection{Moving Average Crossover (50/100-Day, Daily Rebalance)}
A trend-following rule that holds a ticker only when its 50-day simple moving average is above its 100-day simple moving average. Positions are rebalanced daily: capital is equally allocated to all tickers meeting the condition (cash otherwise). This classic dual-moving-average system aims to ride trends while avoiding major drawdowns.

\subsubsection{Random Noise (Control)}
A deliberately naive control strategy that, each day, applies random weights (±40\% long, ±40\% short, 20\% zero per ticker, normalized) to daily ticker returns. This ``monkey with a dartboard'' baseline helps gauge whether observed performance exceeds what pure randomness could achieve in the same universe.

\subsection{LLM-based Trading Agents}
\label{subsec:ai-agents}

We implement the LLM-based trading agents using the open-source AI Hedge Fund framework~\citep{virattt2024hedge}, a popular proof-of-concept system (45.3k stars on GitHub as of January 2026) for exploring AI-driven investment decisions. The repository is available at \url{https://github.com/virattt/ai-hedge-fund}.

The framework employs a multi-agent architecture where specialized LLM-powered agents---including a Valuation Agent, Sentiment Agent, Fundamentals Agent, Technicals Agent, and others modeled after prominent investors (e.g., Warren Buffett)---analyze market data and generate trading signals. A Risk Manager agent computes position limits based on risk metrics, and a Portfolio Manager agent synthesizes the signals into final allocation decisions. Trades are simulated only (no live execution), making it ideal for research and benchmarking.

Key features relevant to our evaluation include:
\begin{itemize}
\item Support for multiple LLM providers via API (OpenAI, Groq, Anthropic, DeepSeek)
\item Built-in access to historical price data for popular tickers (AAPL, GOOGL, MSFT, NVDA, TSLA) without requiring an external API key---aligning precisely with our trading universe.
\end{itemize}

In our experiments, we adapt the framework to a consistent monthly rebalancing protocol (aligned with the quantitative baselines): at the start of each month, the agents are prompted with historical data up to that point, generate target weights for the five stocks, and the weights are normalized and executed (fractional shares allowed, equal initial capital).

We evaluate the following models within this agentic workflow:

\paragraph{Standard Foundation Models}
Widely used open-source LLMs with potential training data contamination extending into 2023--2024:
\begin{itemize}
\item Meta Llama~3.1 (8B and 70B parameters), with cutoff date of December 2023
\item DeepSeek~3.2, with cutoff date of July 2024
\end{itemize}

\paragraph{Point-in-Time (PiT) Models}
The proprietary Pitinf family from PiT-Inference~\citep{pitinference2026}, designed with effective temporal cutoffs of January 2020 to eliminate lookahead bias:
\begin{itemize}
\item Pitinf-Small ($<$10B parameters) --- suited for low-latency tasks.
\item Pitinf-Medium ($<$100B parameters).
\item Pitinf-Large ($>$500B parameters) --- for frontier-level reasoning performance.
\end{itemize}

\section{Experimental Results}
\label{sec:results}

Table~\ref{tab:results} summarizes the performance of quantitative baselines and AI models across both periods.

\begin{table*}[t]
\caption{Performance Comparison across quantitative baselines and AI models. Note the "Scaling Paradox" in P2: while Standard models degrade as they scale (due to stronger false priors), PiT models improve as they scale (due to cleaner reasoning).}
\label{tab:results}
\centering
\small
\begin{tabular}{@{}llrrrrr@{}}
\toprule
\textbf{Model / Strategy} & \textbf{Variant} & \textbf{P1 Return} & \textbf{P1 Alpha} & \textbf{P2 Return} & \textbf{P2 Alpha} & \textbf{Alpha Decay} \\
& & \textbf{(\%)} & \textbf{(pp)} & \textbf{(\%)} & \textbf{(pp)} \\
\midrule
\multicolumn{7}{l}{\textit{QUANT STRATEGIES (Baseline)}} \\
\midrule
Buy \& Hold & Passive & +25.32 & +0.00 & +24.75 & +0.00 & +0.00 \\
Equal Weight & Systematic & +25.68 & +0.36 & +22.33 & -2.42 & -2.78 \\
Momentum (3M) & Systematic & +33.28 & +7.96 & +30.50 & +5.75 & -2.21 \\
Mean Reversion & Systematic & +20.70 & -4.62 & +9.35 & -15.40 & -10.78 \\
MA Crossover & Trend & -2.46 & -27.78 & +6.91 & -17.84 & +9.94 \\
Random Noise & Control & +11.43 & -13.89 & +0.56 & -24.19 & -10.30 \\
\midrule
\multicolumn{7}{l}{\textit{AI MODELS (Agents)}} \\
\midrule
Llama 3.1 8B & Standard & +39.13 & +13.81 & +21.33 & -3.42 & -17.23 \\
Llama 3.1 70B & Standard & +44.59 & +19.27 & +28.77 & +4.02 & -15.25 \\
DeepSeek 3.2 & Standard & +46.05 & +20.73 & +23.71 & -1.04 & -21.77 \\
Pitinf-Small & PiT & +25.07 & -0.25 & +24.81 & +0.06 & +0.31 \\
Pitinf-Medium & PiT & +27.76 & +2.44 & +28.04 & +3.29 & +0.85 \\
Pitinf-Large & PiT & +31.34 & +6.02 & +32.07 & +7.32 & +1.30 \\
\bottomrule
\end{tabular}
\end{table*}

\subsection{Key Insights}

\paragraph{Relative to Quant Strategies:}

\begin{itemize}
\item \textbf{Momentum Robustness:} The best quantitative strategy (Momentum) generated +7.96pp alpha in P1 and maintained +5.75pp in P2, showing only mild decay (-2.21pp).

\item \textbf{Mean Reversion Failure:} The worst strategy (Mean Reversion) decayed significantly (-10.78pp), highlighting the risk of overfitting strategies to specific mean-reverting regimes in trending markets.
\end{itemize}

\paragraph{The Scaling Paradox and Inverse Scaling:}

Our results highlight a critical divergence in how model scaling affects performance in the presence of lookahead bias. We observe what we term the "Scaling Paradox":

\begin{itemize}
\item \textbf{Standard Models (The "Memory Trap"):} For standard models, scaling size actually \textit{hurt} P2 performance relative to the peak. While DeepSeek 3.2 achieved the highest P1 Alpha (+20.73pp) due to superior memorization of 2021 data, it suffered the most severe collapse in P2 (-21.77pp decay).

\item \textit{Interpretation:} This aligns with the "Inverse Scaling" phenomenon \citep{mckenzie2023inverse}, where performance on specific tasks degrades as model scale increases. In financial forecasting, larger models develop stronger, more brittle priors based on training data. As noted by \citet{lopezlira2025memorization}, larger models are more adept at recalling exact historical values (perfect memory), but this "photographic" recall becomes a liability during regime shifts. When deployed in P2 (2024), these strong internal priors conflict with reality, leading to confident hallucinations that override immediate contextual signals.

\item \textbf{Point-in-Time Models (The "Reasoning Dividend"):} Conversely, the Pitinf family demonstrates a positive scaling law in the absence of bias. Once the distractor of "future memory" is removed, scaling the model size improves genuine financial reasoning—such as sentiment interpretation and macro-analysis—rather than merely scaling the capacity to memorize historical price paths. 
\end{itemize}

\section{Conclusion and Future Work}
\label{sec:conclusion}

This paper is a proof-of-concept of how Look-Ahead-Bench can serve as a diagnostic tool for look-ahead bias in financial LLMs. Standard foundation models, while powerful, suffer from severe alpha decay when moved from memorized training windows to new data. In contrast, Pitinf models demonstrate better generalization.

However, more experimentations would be needed, such as:

\subsection{Expand Experimental Scope}

The current evaluation is restricted to five large-cap US technology stocks. To ensure robustness, more diverse backtesting settings are needed, with at least 20--30 stocks across multiple sectors, including:
\begin{itemize}
\item Small-cap stocks: These are less likely to be memorized in training corpora.
\item Non-tech sectors: E.g., Healthcare, Industrials.
\item Diverse Assets: Commodities or FX, though LLM applicability there differs.
\item Multiple Time Periods: Testing 4+ periods of varying lengths.
\end{itemize}

\subsection{Expand Models and Agents}

Look-ahead Bench needs to be enriched with a broader range of trading agents, such as FinMem \citep{yu2023finmem}, FinGPT \citep{yang2023fingpt}, FinRL-DeepSeek \citep{benhenda2025finrl}, TradingAgents \citep{xiao2025trading}, and Hedge-Agents \citep{li2025hedge}. 

Additionally, the scope of autonomous trading architectures should be expanded by integrating specialized agents including FinAgent \citep{zhang2024finagent}, FinRobot \citep{huang2024finrobot}, and StockAgent \citep{ming2024stockagent}. Furthermore, QuantAgent \citep{li2025quantagent} needs to be incorporated for quantitative investment strategies, while CryptoTrade \citep{zhao2024cryptotrade} should be added to extend coverage to cryptocurrency markets.

To assess fundamental analysis and predictive capabilities, models focusing on factor discovery and market forecasting should be considered. This should include LLMFactor \citep{li2024llmfactor} for quantitative factor mining, as well as predictive agents such as TradingGPT \citep{li2023tradinggpt}, AlphaGPT \citep{li2023alphagpt}, MarketGPT \citep{marketgpt2024}, and StockGPT \citep{stockgpt2024}. Evaluation frameworks like PIXIU \citep{xie2023pixiu} and FinPos \citep{liubijia2025finpos} should also be utilized for their comprehensive datasets and insights into position sizing.

Finally, to address complex market dynamics involving interaction and high-frequency trading, the benchmark needs to be extended to support multi-agent reinforcement learning environments. This should involve frameworks such as Language Model Guided RL \citep{llmrl2025}. Recent tools like AutoTrader \citep{autotrader2025} should also be examined to cover the spectrum from research prototypes to deployable trading systems.

\subsection{Expand Backtesting Methodologies}

The benchmark should aim to move beyond historical price paths by employing advanced validation techniques, including:
\begin{itemize}
\item Synthetic Data and Counterfactuals: Testing how models react to market events that could have happened but didn't.
\item Rademacher Anti-Serum (RAS): Utilizing methods like those proposed by \citet{paleologo2025elements} to rigorously test for backtesting overfitting.
\end{itemize}

\bibliography{references_lookahead}
\bibliographystyle{icml2026}

\end{document}